# Participatory Research as a Path to Community-Informed, Gender-Fair Machine Translation

**Dagmar Gromann**[1], **Manuel Lardelli**[2], **Katta Spiel**[3], **Sabrina Burtscher**[3], **Lukas Daniel Klausner**[4],
**Arthur Mettinger**[5], **Igor Miladinovic**[5], **Sigrid Schefer-Wenzl**[5], **Daniela Duh**[5], and **Katharina Bühn**[5]

[1]University of Vienna, Austria, `{name.surname}@univie.ac.at`
[2]University of Graz, Austria, `{name.surname}@uni-graz.at`
[3]TU Wien, Austria, `{name.surname}@tuwien.ac.at`
[4]St. Pölten University of Applied Sciences, Austria, `mail@l17r.eu`
[5]FH Campus Wien University of Applied Sciences
`{name.surname}@fh-campuswien.ac.at`

## Abstract

Recent years have seen a strongly increased visibility of non-binary people in public discourse. Accordingly, considerations of gender-fair language go beyond a binary conception of male/female. However, language technology, especially machine translation (MT), still suffers from binary gender bias. Proposing a solution for gender-fair MT beyond the binary from a purely technological perspective might fall short to accommodate different target user groups and in the worst case might lead to misgendering. To address this challenge, we propose a method and case study building on participatory action research to include experiential experts, i.e., queer and non-binary people, translators, and MT experts, in the MT design process. The case study focuses on German, where central findings are the importance of context dependency to avoid identity invalidation and a desire for customizable MT solutions.

## 1 Introduction

With an increased visibility of non-binary people in public discourse, gender-fair language strategies to go beyond a binary conception of male/female have been proposed. Gender-fair language subsumes gender-inclusive, i.e., linguistically including all gender identities, and gender-neutral, i.e., removing all gender references, strategies. Practically applying gender-fair language across grammatically different languages is challenging for human and machine translation. In human gender-fair translation, substantial errors can be observed (Lardelli and Gromann, Forthcoming; Attig, 2022). In MT, the "masculine default" might have been mitigated with strategies to debias MT, however, generally with a binary focus (Savoldi et al., 2021) and not linguistically acknowledging existing gender identities with few exceptions (Saunders and Byrne, 2020; Piergentili et al., 2023). Savoldi et al. (2021) call for research beyond NLP and its "narrow, problem-solving oriented approach" to advance the field and Attig (2022) proposes to include queer and non-binary people in a community-informed translation process.

The proposed case study builds on the notion that gender-fair (machine) translation requires an early community involvement. The word "machine" is at times placed in brackets since there was a common consensus that first gender-fair translation strategies are required to facilitate gender-fair MT. To this end, ten researchers in Austria organized a three-day workshop with in total 21 participants from three groups of stakeholders, i.e., queer and non-binary people, professional translators, and MT experts, to reflect on their experiences, desires, and concerns regarding MT. Furthermore, we seek to provide a method that emphasizes the importance of human value, similar to the Diverse Voices method (Young et al., 2019), and includes marginalized groups in technology design, i.e., with participatory design (Spiel et al., 2020). In the proposed method, Participatory Action Research (PAR) is utilized to design a set of activities to identify problems, desires, strategies, and proposed adaptations to the MT design process, as depicted in Fig. 1.

In MT, it has been proposed to solve the issue by focusing on gender-neutral strategies (Piergentili et al., 2023), which, however, cannot be applied to all contexts, especially not without information loss, and might not be the strategy preferred by all MT users. In fact, findings from this workshop challenge the current MT paradigm of one input equals one output and reveal strong preferences for a customizable solutions that allow users to select their preferred strategy and make context-informed suggestions. To the best of our knowledge, this is the first community-informed workshop on gender-fair (machine) translation with the explicit purpose to include human value and marginalized groups in the technology design process. In this paper, we describe the activities themselves, their practical implementation as a workshop with three groups of stakeholders, and a potential blueprint of the activities to develop similar workshops for other languages and communities.

## 2 Preliminaries

To provide a theoretical basis for the following discussion, we first briefly explain the discourse on gender and language, and PAR as well as stakeholder selection taken from value-sensitive design.

### 2.1 Gender and Language

The term gender involves biological aspects, i.e., functionality of brains, production of hormones, as well as psychological, i.e., the way people identify, experience, think about gender, and social aspects, i.e., how gender is enacted in a particular context (Barker and Iantaffi, 2019). Gender identities are self-determined and not assigned (Zimman, 2019) and beyond the male/female dichotomy range from agender, genderfluid to non-binary or pangender among many others (Richards et al., 2016). Gender identities are crucial in daily lives since they are used as a criterion to regulate access to services and goods (Fae, 2016) as well as public spaces, such as restrooms (van Anders et al., 2017).

Structurally, natural languages have been categorized into grammatical gender, notional gender, and genderless languages (Stahlberg et al., 2007; Savoldi et al., 2021). In grammatical gender languages, nouns, adjectives, pronouns, and determiners are gender-inflected. In notional gender languages, such as English, lexical gender, such as *boy* and *girl*, derivational nouns, such as *waiter* and *waitress*, and pronouns are gender-specific. In genderless languages, such as Finnish, mostly references to kinship are gendered, e.g. *sister* and *brother*. Gender is not only a matter of grammatical gender, but also a social construct and can emerge in the way language is used, which reflects assumptions and norms on gender identity.

**Gender-Fair German** Inspired by Sczesny et al. (2016), we subsume gender-inclusive and gender-neutral strategies as gender-fair language. Gender-inclusive strategies, which seek to make all genders visible, can be of two different types: (1) typographical characters (*, :, _) to separate male from female forms and include all genders, e.g. *Leser\*innen* (reader) and *sie\*er* (she*he) (Hornscheidt and Sammla, 2021); (2) new gender systems, such as the SYLVAIN system (de Sylvain and Balzer, 2008), that introduce a fourth grammatical gender in addition to masculine, feminine and neuter, e.g. *Lesernin* (reader). Gender-neutral strategies can vary, ranging from the use of typographical characters to remove any gender endings, such as *Les\**, and rewording to the introduction of neutral endings and pronouns, e.g. *Lesens* and *ens* (Hornscheidt and Sammla, 2021).

**Impact of Gender Bias** Bias in language technology can be defined as systems that "*systematically* and *unfairly discriminate* against certain individuals or groups of individuals in favour of others" (Friedman and Nissenbaum, 1996). Bias in the training data leads to MT systems with biased predictions, e.g. sampling non-random subsets of data. The main impact of gender bias is the harm that can be produced by a biased system. Crawford (2017) differentiates between allocational and representational harms. The former refer to the allocation or withholding of opportunities or resources to certain groups. The latter refer to lessening or omitting the representation of specific groups and their identity. Not recognizing the existence of gender beyond the binary can imply the harm of disregarding the language used by these communities (Savoldi et al., 2021). Misgendering, the assignment of a wrong gender to a person, should be added to the list, which can lead to emotional pain and a feeling of identity invalidation (Zimman, 2019). Finally, stereotyping refers to propagating negative generalizations of a social group (Savoldi et al., 2021).

## 2.2 Participatory Action Research

Participatory Action Research (PAR) is a highly community-led approach that allows for different stances that, when combined, deliver a more vibrant description of agendas and contexts of use than researchers' perspectives could provide on their own (Hayes, 2011). With action research as a methodological base, this method is committed to have a positive transformative impact on the individuals' lives as well as further representatives of the associated communities. Concretely, this means also to actively facilitate negotiation between different stakeholder groups and their potentially contradicting needs and desires.

A typical PAR process consists of alternating action and reflection phases (Kindon et al., 2007), where action focuses on building relationships and performing collaborative activities and reflections focus on research design/process, ethics, knowledge and accountability as well as towards the end on how well the collaboration has worked and further steps that are required. Typical activities include dialog, storytelling, and collective action that with their hands-on nature are particularly adequate for work with marginalized or vulnerable people since they allow participants to generate information and share knowledge on their own terms using their own language (Kindon et al., 2007).

## 2.3 Value Sensitive Design

Value Sensitive Design (VSD) seeks to account for human values in technology design, where values substantially depend on the interests and desires of human beings within a specific context (Friedman et al., 2013). VSD considers direct and indirect stakeholders, where the former are those that directly interact with the technology and the latter are those impacted by the technology, even if potentially never touching the technology itself. Stakeholders may span more than one role in a design process, e.g. translator and non-binary in our case study. VSD can further help in deciding on which stakeholder groups to prioritize, where we follow Young et al. (2019) in prioritizing transparency over specific ethical or other concerns. We seek to include these groups that impact and are most likely impacted by the technology, but who have yet been underrepresented in previous research, i.e., queer and non-binary people.

## 3 Objective

In our initial endeavors to develop a gender-fair MT model, we quickly realized that we not only lack training datasets but sufficient knowledge to decide on a gender-fair strategy for German. Thus, our objective in organizing a workshop was to bring together three communities to jointly discuss considerations and implications that should be taken into account in designing gender-fair MT solutions.

Queer and non-binary people represent indirect stakeholders in the sense that their interaction with MT in the past or future is neither a given nor a requirement. Nevertheless, this group is substantially impacted by gender-fair language use or the lack thereof. The active use of gender-inclusive or gender-neutral strategies in MT could positively affect the visibility of gender diversity, since its use is widespread and users are frequently not even aware of consuming MT outputs (Martindale and Carpuat, 2018).

Professional translators are both direct and indirect stakeholders. In spite of some resistance (Cadwell et al., 2018), MT is increasingly integrated into translation pipelines (Way, 2020), which makes translators direct stakeholders. On the other hand, as providers of translations for training MT models, they are indirect stakeholders who impact the technology. While translators at times might have been included in post-editing or MT evaluation studies, the two groups above have to the best of our knowledge not been included in the MT design and development process.

Finally, we decided to select MT developers/researchers as a stakeholder group to gather insights into the feasibility of ideas devised in group discussions and include their perspective on this topic in the cross-community exchange. To reach stakeholders, it is vital to include facilitators who are trustworthy to the respective communities, in particular in case of sensitive topics, and who can rely on personal networks and contacts for invitations of stakeholders. Furthermore, such facilitators, in our case as part of the research team, need to actively help shape the plan in an appropriate and acceptable manner for participants, moderate the workshop, and intervene in group discussions if stagnation or conflicts arise.

In terms of method, our objective was to encourage participants to share their experiences, desires, and concerns as well as informed critique on ex-isting gender-fair (machine) translation strategies and requirements. Participatory Action Research (PAR) is designed to elicit and generate situated knowledge that provides insights beyond a unique case study and ensures an interactive, motivating design. Thus, the design of our activities follows the PAR principles and cycles.

## 4 Participatory Workshop Activities

The workshop plan was designed to alternate interactive sessions in small groups and plenary discussions on each of the five main topics and stages depicted in Fig. 1. Activities in small groups deliberately alternated between groups constituted by members of the same stakeholder group (community-internal) and groups with members of each stakeholder group (cross-community). Each interactive group activity was planned for approx. one hour and accompanied by written instructions, followed by approx. 30 minutes of presentation of results and plenary discussion. At the end of each day, a joint summary of major topics and findings was to be prepared in plenary session, including interactive quizzes and word clouds on Mentimeter.

### 4.1 Warm-Up Phase

For the registration, voluntary colored stickers to indicate a) the stakeholder group and b) the pronouns of participants were foreseen. As an icebreaker, a first sociometric introduction asked participants to position themselves along different axes of the room for a number of off-topic, e.g. means of transportation to arrive at the workshop, and on-topic questions, e.g. familiarity with gender-fair language and MT, to establish relations among participants.

### 4.2 Stage 1: Problem Storming

A first problem storming session in a cross-community setting of three to four participants per group, where each community was represented, targeted an exchange of experiences, interests, and needs as well as potential challenges in reference to gender-fair (machine) translation. To initiate the discussion, different text samples were provided: descriptions of non-binary people (Example (1)), mixed-gender groups (Example (2)), and without gender indication (Example (3)). For the last two, we provided the English source text with "they" for mixed-gender groups and two MT outputs in German produced with Google Translate and DeepL. The use of colored cards on a pin board was suggested to analyze issues in the (machine) translation of such texts.

(1) Eliot Sumner ist Musiker*in, Schauspieler*in und als Kind von Sting und Trudie Styler quasi im Entertainment-Biz aufgewachsen. Außerdem ist Eliot nichtbinär, identifiziert sich also weder als Mann noch als Frau, weshalb wir hier das nichtbinäre Pronomen "xier" benutzen.

(2) Did someone leave their books here?
    a. Hat jemand seine Bücher hier liegen lassen? (Google Translate & DeepL)

(3) An employee will not do a good job if they don't have the right training.
    a. Ein Mitarbeiter wird keinen guten Job machen, wenn er nicht die richtige Ausbildung hat. (Google Translate)
    b. Ein Mitarbeiter wird keine gute Arbeit leisten, wenn er nicht die richtige Ausbildung hat. (DeepL)

### 4.3 Stage 2: Utopia Storming

The second stage represented a community-internal group activity and instructs each stakeholder group to jointly dream up a social and technological utopia, where the focus was explicitly not on feasibility but on dreams, hopes, and desires. Here all materials available in the room could be used, including wool, pipe cleaners, etc. At the end of Stage 2, the day concluded with a summary sessions and two Mentimeter word clouds, one on the greatest insights and a second one on the greatest barriers for gender-fair (machine) translation.

### 4.4 Stage 3: Hands-On

To not go directly from utopias to strategic considerations on the second day, we intercepted the process with a hands-on stage where specific examples of use can be analyzed and the preparatory handout can be put to practice. Cross-community groups of three to four participants obtained profiles of fictional characters, such as Ariel, The Little Mermaid, Peter Pan, Pippi Longstocking (see Fig. 2), with the task to prepare their introduction in gender-fair language. The objective of the activity was to raise awareness on the degree of gender specificity in German, one's own language use, and the multiciplity of gender-fair strategies, of which each group was instructed to select one.

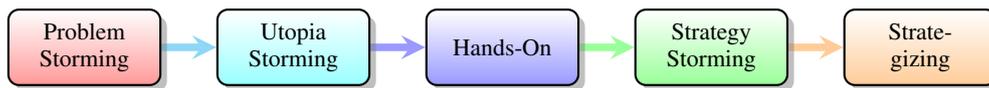

Figure 1: Stages for Participatory Workshop on Technology Design

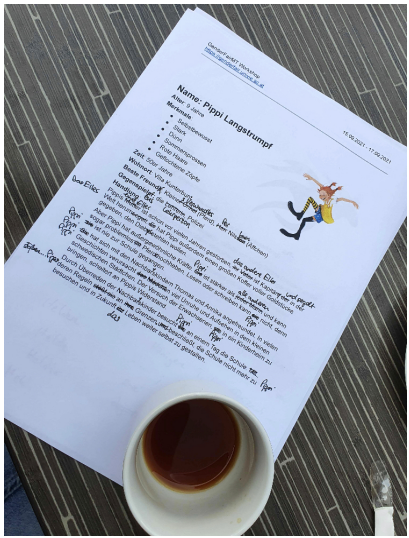

Figure 2: Presenting fictional characters as non-binary

### 4.5 Stage 4: Strategy Storming

Having identified problems and a desired utopia, the strategy storming stage seeks to gradually turn towards the concrete and potential approaches in an open process endorsing creativity rather than feasibility, which was explicitly described as the focus of the next stage. For this stage, participants were asked to visualize their results, e.g. on flipcharts or with colorful sticky notes.

### 4.6 Stage 5: Strategizing

Finally, strategizing focuses on the social as well as technical feasibility of discussed strategies. The initial session was a cross-community task within the same grouping as in the strategy storming session. The last day was dedicated to first community-internal sessions on potential cross-fertilization initiatives across communities, followed by a cross-community session on who needs what from whom. At the end of the second day, we asked participants for their preferred gender-fair strategy by means of Mentimeter quizzes.

### 4.7 Synthesizing

At the end of the third day, a summary of the most central insights and implications was jointly prepared in a plenary session. For a summary at the end of each day as well as for this final summary, we recommend taking notes online live and projecting these notes so that participants can correct potential mistakes directly. Furthermore, the final summary is circulated to participants for inspection, expansion, and correction. To avoid losing the momentum of a successful community building effort and event, concrete steps to interact beyond a sharing of workshop outcomes and further research endeavors should be taken. Following the principles of PAR, a democratic and self-determined method should be foreseen, such as a mailing list, a Wiki, or any other means of interchange of ideas, which at best should be decided together with the participants.

## 5 Participatory Workshop in Action

A team of ten researchers from Austria conducted a three-day participatory workshop on the topic of gender-fair MT with an initial focus of translating from and to German. Our research team consisted of members of and people closely connected to the queer and non-binary community, professional translators with good contacts in this community, active MT and human-computer interaction researchers with a corresponding network. Thus, we could strongly rely on our personal networks to invite participants and instill the necessary trust for participants to accept the invitation. Participants were additionally recruited through activist groups and open calls, following a sampling strategy that allows for a spread of different marginalized experiences including intersecting aspects of marginalization.

The workshop initially targeted ten participants for each community and stakeholder group. Finally, in total 21 people participated, ten translators, six MT experts, and five queer and non-binary people. Several of these 21 participants had intersectional roles, e.g. translator and MT expert or non-binary and translator. Given that the workshop was organized amidst the pandemic, we were grateful for this turnout and the substantial commitment of participants to take three days off their working week, one MT expert even traveled from Switzerland. To establish a common basis of knowledge, we distributed a preparatory handout summarizing several gender-fair language strategies for German[1]. Additionally, the handout provided pointers for further reading.

In the following we present the results from the workshop activities (see Section 4) by identified problems, proposed utopias, and strategic considerations for realizing gender-fair (machine) translation.

## 5.1 Problems

In terms of barriers to gender-fair (machine) translation, most participants indicated acceptance, ignorance, lack of resources, and a lack of understanding. One central issue that was identified across stakeholder groups is the linguistic creativity of the German language and the respective higher effort to effect change to achieve gender-fair language use. Ideally, people should be directly asked for their identified pronouns and language strategy, which, however, is practically not feasible for written texts in MT or mixed groups. Participants agreed that numerous linguistic aspects are strongly context-dependent and one general gender-fair language strategy will hardly accommodate all possible contexts. In reference to MT specifically, a lack of gender-fair text samples and training corpora was addressed. Already at this stage first ideas towards solutions were proposed, e.g. to enable users to select the desired gender-fair strategy in the target text and to potentially implement translations from German to gender-fair German as a first step. One central issue unanimously agreed upon is that language alone will not suffice to achieve inclusivity if only linguistic surface forms are changed, but stereotypical gender ideas and reactionary thinking remain unaltered.

## 5.2 Utopias

Utopia storming in a community-internal setting targeted hopes and dreams of a better, gender-fair world supported by technology, where all available materials could be used. Fig. 3 exemplifies the creativity of the groups, i.e., the MT lego unicorn (a) and the translators' "Eierlegendes Wollmilch Ich-bin-Ich" (egg-laying wool-milk I-Am-Me)[2] (b).

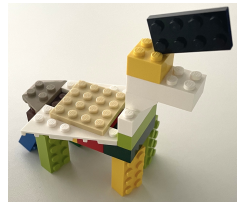 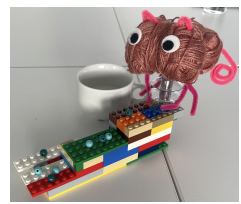

**(a)** "Leonda, the Gender Avenger"  **(b)** "Eierlegendes Wollmilch Ich-bin-Ich"

**Figure 3:** Shared visualized utopias

The utopia of translators and MT experts advocated guidelines and standards to allow for an easier practical and technical implementation. Standardization could foster acceptance of gender-fair language and simplify language patterns, which could be beneficial to MT. Nevertheless, any standard should be flexible enough to allow for visibility of people disregarded by its necessary reduction. Another interesting idea was a point of contact for gender-fair language, such as a helpline. The utopia of the queer and non-binary group indicated rather a harsh reality, since they desired respect, sensitivity and "just to be able to exist". In the following joint plenary session the necessity of political, legal and social frameworks were discussed as a means to foster the demand for gender-fair language, e.g. to provide incentives for enterprises to achieve specific gender-fair language goals, such as a gender-fair certification mark.

## 5.3 Strategies

When directly asked about the preferred gender-fair language strategy on the second day, exactly half of voting participants (n=20) preferred gender-inclusive and the other half gender-neutral language. In the hands-on Stage 3, a predominant strategy consisted in omitting pronouns, utilizing names and passive constructions, and substituting nouns with plural or neutral variations, e.g. *Meerjungfrau* (mermaid) became *Meerwesen* (merbeing). Even though for several participants omitting pronouns seemed easier than using neopronouns, e.g. *nin*, this inevitably led to a more frequent repetition of names and subjectively less frequent sentence structures. Two groups utilized the gender-inclusive Dey-E-System[3]. Participants familiar with gender-fair language use found writ-ing texts from scratch easier than 'translating' an existing profile to gender-fair language, with the argument that gender-specific elements can easily be overlooked. Such involuntary omission is not only a challenge for human beings, but equally for MT, since designing a system that detects infrequent or less obvious mentions, such as *mermaid*, as gender-specific and is capable to provide a gender-fair alternative is definitely an open research issue.

---

[1] https://genderfairmt.univie.ac.at/files/Handout_Genderfaires_Deutsch.pdf
[2] It reflects on the "eierlegende Wollmilchsau", a colloquialism to indicate something that can cater to all needs exemplified in the unreal animal providing eggs, wool, and milk, and "Das kleine Ich bin ich" (Little I-A-Me) is a reference to a children's book.
[3] As in *einey gute Ärzte* (a good doctor); see https://geschlechtsneutral.net/dey-e-system/

When asked to consciously select a strategy, each group preferred a different solution: (i) a multi-stage model, (ii) the gender-neutral ens strategy, and (iii) the gender-inclusive SYLVAIN system. The idea of a multi-stage model was to clearly assess one's own language use and gradually progress towards gender-fair language. The bottom stage linguistically includes women, e.g. utilizing male and female forms, the German Binnen-I (*LeserInnen*), and only using female forms (*Leserinnen*). The second stage includes non-binary people by utilizing gender-inclusive characters (*, :, _), where MT experts remarked on the issue of * being a syntactic element of technical languages, e.g. to represent text in italics. The final stage aims to avoid outing individuals by linguistic means, which can be achieved with gender-neutral strategies. This idea of a multi-stage model towards gender-fair language could be implemented as a multi-stage MT adaptation process.

The second group preferred the *-ens* strategy as in M**ens**ch (human being), e.g. *Lesens* (reader). Utilizing this strategy would resolve the issue of special characters, character and text length, and pronunciation and readability. Arguments against this strategy were that this form is too similar to the genitive case in German, which might lead to confusions, and that no distinction between singular and plural is foreseen, which leads to further omission of information apart from gender-specific omissions.

The third group preferred the SYLVAIN system that introduces a new gender, the liminal gender, due to a preference of inclusion over omission of gender and with the arguments that contexts can be preserved, direct translation equivalents are facilitated, and a consistent use of language is eased without omissions of information. Nevertheless, translating gender-neutral elements with the SYLVAIN system, e.g. English singular *they* to *nin*, would change the context of the source text and might erroneously assign a liminal gender.

## 5.4 Strategizing

One suggested solution to overcome the disparity between gender-inclusive and gender-neutral forms was to develop a hybrid form that uses gender-neutral forms and simultaneously permits gender-specific references. However, the criterion of not involuntarily outing individuals might still be an issue in such a hybrid model. In addition to this criterion, practicability, ease of access and pronunciation, universality and acceptability were proposed. To ensure inclusion of diverse groups, including language learners and people with disability, comprehensibility and readability should be taken into consideration. For instance, *Lesx* or *Les\** represent gender-neutral language but are neither straightforward to pronounce, comprehend, or apply for first language speakers of German.

From a business perspective, it was deemed essential to achieve and ensure a consistent use of language, e.g. for search engine optimization, whether when writing new contents or translating existing ones. This brought up the idea of standards or guidelines again, which could increase the confidence in grammatical correctness and unify pronunciation. Furthermore, a guideline would ease adoption and support from a social and societal standpoint and equally from an institutional perspective, e.g. major dictionaries of the German language as the Duden, media, or public authorities. Referencing and addressing unknown people would considerably be eased by such standardization, however, such an endeavor is in opposition to the dynamically evolving language and gender-fair language, where strategies are still developing within the queer and non-binary community. As an alternative, flexible guidelines potentially combine a degree of standardization with open possibilities to personalize language.

A reduction of the multiplicity of gender-fair language strategies could ease the generation and availability of gender-fair texts and training data to facilitate MT. In this context, the idea of rule-based generation of text samples and a novel professional profile of a community-based gender-fair pre- and post-editor were discussed. Whether rule-based or implemented differently, MT systems were seen as central tools to explore different approaches to gender-fair language systems.

## 5.5 Cross-Community Support

In the discussion of mutual support across communities, the translation community suggested jointly creating a cheat sheet for gender-fair language that can be consulted during the translation process, official guidelines to justify translation decisions for clients, training workshops from the queer and non-binary community, and tools to facilitate gender-fair translation. The queer and non-binary community mainly desired gender-fair translations, whether automatically or manually created, to increase gender-fair language use and active and continuous exchange with the other communities, as initiated by this workshop. This community emphasized the role of translators to potentially bridge a gap and facilitate exchange between the majority society and the queer and non-binary community. Furthermore, the community often feels like applicants or solicitors to be included and thus, would desire to be better included and considered by the other communities. The MT community desired mainly gender-fair text samples and corpora and equally a continuous exchange with the other two communities. A continuous involvement of the other communities in the further progress of a gender-fair MT development process was envisioned. In short, communities were united by the desire for interdisciplinary, "multiprofessional" teamwork to jointly work towards the defined objectives. A very nice visual summary that resulted from a final cross-community group session is depicted in Fig. 4.

## 6 Reusability of the Participatory Workshop

One option to address a more diversified pool of participants and reach a wider as well as slightly bigger audience could be to move a considerably shortened version of these activities online, as e.g. done by Pannitto et al. (2021) with more tool support for interactive sessions, e.g. Miro to replace flipcharts, breakout groups in video conference tools, etc. Since the nature of PAR projects is to be situated in a particular context and relationships in order to generate situated knowledge, targeting large audiences might benefit from a different methodological choice. Nevertheless, a PAR project provides insights with wider implications from unique use cases, called "communicative generalization" (Cornish, 2020). It addresses the "the significance of knowledge to epistemic communities rather than abstract universal truth" (Cornish, 2020), facilitating the expression and perception of multiple perspectives, enriching the reader's generalized other, and problematizing situations that are taken for granted.

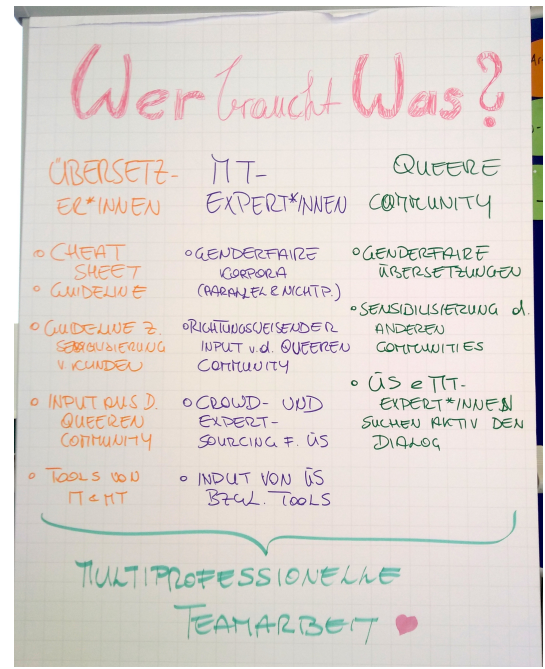

**Figure 4:** Multi-professional teamwork

Our case study focused on the issue of gender-fair (machine) translation from English to German as a starting point. While the proposed English materials might be reusable for workshops focused on other languages, adapting the activities to other languages should take culture- and community-specific considerations into account. Nevertheless, we believe that the general method and structure of the workshop can be utilized as a blueprint for further such workshops. On a general note, we recommend alternating group activities, within and across groups of stakeholders, with plenary sessions for joint reflections. Each group should be observed by one team member in a non-participant manner to ensure to take notes on discussions and to intervene should any conflicts or situations of stagnation arise. For plenary sessions, we recommend live summaries that are projected so that participants can directly add/change notes. We provide the core ideas and principles of each activity to facilitate reproducing the workshop in different languages and contexts.

**Pre-Workshop Preparation** For any given research topic that seeks to involve various, frequently distinct groups of stakeholders, it is vi-tal to ensure that all participants share a common body of knowledge. To this end, a preparatory handout explaining the most essential facts can be distributed to all participants prior to the event. This requires preparing ahead to permit sufficient time for each participant to familiarize themselves with the provided contents. The advantage of this preparatory step is that the workshop can commence with interactive sessions instead of presenting content to participants and participants have the chance to familiarize themselves with the topic at their own pace. Furthermore, it permits time for additional reading in case a single stakeholder feels they would like to know more about a topic – corresponding pointers should be included in the handout.

**Warm-Up** With sensitive topics, such as gender identity in language, we recommend warm-up activities, such as a sociometric introduction (see Section 4.1), and subtle as well as optional means to identify with a specific group of stakeholders or community.

**Stage 1 & 2** At the beginning of the workshop activities, we utilize gender-fair texts and their machine translations to initiate the discussion on potential problems of the topic. These can easily be reproduced for other languages and with any available MT system, since the focus is not on translation adequacy but on triggering a discussion on gender-fair language. Also utopia storming is an easily transferable activity that is best organized offline with a large pool of very different (handicraft) materials, e.g. colorful papers, scissors, building blocks, etc. For this stage, it is important to emphasize that proposed utopias neither need to be possible nor feasible, but could represent any dream vision.

**Stage 3** For the hands-on activity, materials need to be adapted to the respective target culture and community. The chosen fictional characters and corresponding profiles should be well known by the participants to be able to present them in gender-fair language, e.g. there might be a better choice than Pippi Longstocking for other settings. For this activity, enough time should be provided for cross-community groups to consider different potential solutions before presenting the one of their choice for the specific fictional character.

**Stage 4** After the practical application of strategies, the goal of strategy storming is to consciously think about preferred strategies for (machine) translation and in general. This stage can directly be transferred by adapting the instructions to the specific language.

**Stage 5** The final stage seeks to discuss feasible solutions and their socio-technical implications as well as mutually beneficial aspects of community exchange. For this stage, we strongly recommend going from a community-internal to a cross-community group session in order to first discuss ideas within groups of stakeholders and then exchange these among groups. For this stage, materials to visualize thoughts and results are also very important to allow for groups to summarize their main points and sort their ideas.

# 7 Key MT Implications

As an overview, we summarize the key implications for gender-fair (machine) translation in the following list:

- need for user-centric, customizable selection of gender-fair language strategy in the target language

- gender-fair MT output(s) depend not only on the input but on the context, people addressed, purpose, and user preferences

- potential need to perform intralingual rewriting, e.g. from German to different gender-fair versions of German

- preference to combine gender-neutral with gender-inclusive language to minimize information loss

- awareness that gender-fair language is language-specific and a quickly evolving field, requiring flexible, adaptable solutions

- general criteria to select a gender-fair language strategy, which entail future (psycholinguistic) research:
    - readability and comprehensibility
    - ensuring not to involuntarily out someone
    - practicability and universality
    - ease of access and pronunciation

## 8 Discussion and Conclusion

As becomes evident from these results, a straightforward decision on a single strategy to ensure linguistic inclusivity is not feasible and this decision should depend on the context – to quote one participant of the workshop "one size fails all". A disparity between a desire to standardize and to personalize gender-fair language brought the discussion to the conclusion that a customizable MT implementation would be most beneficial. It should allow users to flexibly select which strategy to use for a text and, where possible, make informed suggestions for a context-specific strategy.

PAR-based activities gradually brought new arguments from different communities to light and resulted in a catalog of criteria to guide the selection process of gender-fair language strategies for (machine) translation from the multiplicity of dynamically growing proposals, including practicability, ease of access, and universality. Additionally, a central criterion was to provide means of addressing individuals without involuntarily outing their gender identity. Furthermore, any gender-fair language use should be readable, comprehensible, and easy to learn and pronounce. In many cases, gender-neutral strategies, such as *-ens*, comply with these criteria, however, in a translation setting the inherent loss of context-specific information by omitting gender-specific information and plural forms might not be feasible. In a translation setting, a context-preserving target text irrespective of the specific strategy selected was deemed essential as well as further experiments on their translatability.

One central issue with any gender-fair language strategy was a current lack of text samples for training MT systems but also for teaching and exemplifying each strategy, which mostly rely on conjugation and declination tables for their introduction. Hands-on examples clearly showed that any automated method to detect and potentially alter gender-specific mentions in a text needs to go beyond grammatical gender or linguistic surface forms to also detect less obvious examples, such as *mermaid*. To overcome the issue of data for MT, hybrid methods with rule-based elements to synthetically generate text samples were proposed as well as to initiate MT adaptations with a community-informed intralingual system to translate from German to gender-fair German.

In short, this idea of accommodating several gender-fair target texts depending on context and/or user preferences would fundamentally change the current MT paradigm, which relies on the correspondence of one source text with one target text. While the idea of providing different target texts to choose from has entered the world of commercial MT systems, e.g. allowing users to choose between male and female target sentences for a given input, this customization of MT to personalized gender-fair MT target texts would require further substantial adaptations and a community-informed, context-dependent decision on which gender-fair strategies to display if none are indicated by the user. To initiate this development, further research on the selection of gender-fair language strategies that comply with the identified criteria is planned as future work, especially readability and comprehensibility.

As an overall feedback on the workshop, participants were satisfied with the respectful, productive, and constructive atmosphere and there was a general consensus to have gathered new knowledge from the cross-community and community-internal exchanges. Concrete steps for continuing this inter- and transdisciplinary multi-professional teamwork in terms of readability studies and procuring gender-fair text samples have already been initiated. We as organizers were very grateful for the wealth of socio-technical ideas and arguments contributed by participants. We hope that their input will be used to guide future research on gender-fair MT.

## References

Attig, Remy. 2022. A call for community-informed translation: Respecting queer self-determination across linguistic lines. *Translation and Interpreting Studies*.

Barker, Meg John and Alex Iantaffi. 2019. *Life isn't Binary*. Jessica Kingsley Publishers, London, UK.

Cadwell, Patrick, Sharon O'Brien, and Carlos S. C. Teixeira. 2018. Resistance and accommodation: factors for the (non-) adoption of machine translation among professional translators. *Perspectives*, 26(3):301–321.

Cornish, Flora. 2020. Communicative generalisation: Dialogical means of advancing knowledge through a case study of an 'unprecedented' disaster. *Culture & Psychology*, 26(1):78–95.

Crawford, Kate. 2017. The trouble with bias. In *Conference on Neural Information ProcessingSystems (NIPS) – Keynote*, Long Beach, USA.

de Sylvain, Cabala and Carsten Balzer. 2008. Die SYLVAIN-Konventionen – Versuch einer "geschlechtergerechten" Grammatik-Transformation der deutschen Sprache. *Liminalis*, 2008(2):40–53.

Fae, Jane. 2016. Non-gendered pronouns are progress for trans and non-trans people alike. *The Guardian*, 14 Dec.

Friedman, Batya and Helen Nissenbaum. 1996. Bias in computer systems. *ACM Trans. Inf. Syst.*, 14(3):330–347, jul.

Friedman, Batya, Peter H. Kahn, Alan Borning, and Alina Huldtgren. 2013. Value sensitive design and information systems. In Doorn, Neelke, Daan Schuurbiers, Ibo van de Poel, and Michael E. Gorman, editors, *Early engagement and new technologies: Opening up the laboratory*, pages 55–95. Springer Netherlands, Dordrecht.

Hayes, Gillian R. 2011. The relationship of action research to human-computer interaction. *ACM Trans. Comput.-Hum. Interact.*, 18(3), August.

Hornscheidt, Lann and Ja'n Sammla. 2021. *Wie schreibe ich divers? Wie spreche ich gendergerecht?: Ein Praxis-Handbuch zu Gender und Sprache*. w_orten & meer, Insel Hiddensee.

Kindon, Sara, Rachel Pain, and Mike Kesby. 2007. *Participatory action research approaches and methods: Connecting people, participation and place*, volume 22. Routledge, New York, NY.

Lardelli, Manuel and Dagmar Gromann. Forthcoming. Translating Non-Binary Coming-Out Reports: Gender-Fair Language Strategies and Use in News Articles. *The Journal of Specialised Translation*.

Martindale, Marianna J. and Marine Carpuat. 2018. Fluency over adequacy: A pilot study in measuring user trust in imperfect MT. *CoRR*, abs/1802.06041.

Pannitto, Ludovica, Lucia Busso, Claudia Roberta Combei, Lucio Messina, Alessio Miaschi, Gabriele Sarti, and Malvina Nissim. 2021. Teaching nlp with bracelets and restaurant menus: An interactive workshop for italian students. *arXiv preprint arXiv:2104.12422*.

Piergentili, Andrea, Dennis Fucci, Beatrice Savoldi, Luisa Bentivogli, and Matteo Negri. 2023. From inclusive language to gender-neutral machine translation. *CoRR*, abs/2301.10075.

Richards, Christina, Walter Pierre Bouman, Leighton Seal, Meg John Barker, Timo O. Nieder, and Guy T'Sjoen. 2016. Non-binary or genderqueer genders. *International Review of Psychiatry*, 28(1):95–102.

Saunders, Danielle and Bill Byrne. 2020. Reducing Gender Bias in Neural Machine Translation as a Domain Adaptation Problem. In *Proceedings of the 58th Annual Meeting of the Association for Computational Linguistics*, ACL 2020, pages 7724–7736, Stroudsburg, PA. ACL.

Savoldi, Beatrice, Marco Gaido, Luisa Bentivogli, Matteo Negri, and Marco Turchi. 2021. Gender bias in machine translation. *Transactions of the Association for Computational Linguistics*, 9:845–874, 08.

Sczesny, Sabine, Magda Formanowicz, and Franziska Moser. 2016. Can gender-fair language reduce gender stereotyping and discrimination? *Frontiers in Psychology*, 7.

Spiel, Katta, Emeline Brulé, Christopher Frauenberger, Gilles Bailley, and Geraldine Fitzpatrick. 2020. In the details: the micro-ethics of negotiations and in-situ judgements in participatory design with marginalised children. *CoDesign*, 16(1):45–65. PMID: 32406393.

Stahlberg, Dagmar, Friederike Braun, Lisa Irmen, and Sabine Sczesny. 2007. Representation of the Sexes in Language. In Fiedler, Klaus, editor, *Social Communication*, Frontiers of Social Psychology, pages 163–187. Psychology Press, New York, NY.

van Anders, Sari M, Zach C Schudson, Emma C Abed, William J Beischel, Emily R Dibble, Olivia D Gunther, Val J Kutchko, and Elisabeth R Silver. 2017. Biological sex, gender, and public policy. *Policy Insights from the Behavioral and Brain Sciences*, 4(2):194–201.

Way, Andy. 2020. Machine translation: where are we at today? In Angelone, Erik, Maureen Ehrensberger-Dow, and Gary Massey, editors, *The Bloomsbury companion to language industry studies*, pages 311–332. Bloomsbury Publishing Plc, London, UK.

Young, Meg, Lassana Magassa, and Batya Friedman. 2019. Toward inclusive tech policy design: a method for underrepresented voices to strengthen tech policy documents. *Ethics and Information Technology*, 21(2):89–103.

Zimman, Lal. 2019. Trans self-identification and the language of neoliberal selfhood: Agency, power, and the limits of monologic discourse. *International Journal of the Sociology of Language*, 2019(256):147–175.